\documentclass[conference]{IEEEtran}
\IEEEoverridecommandlockouts
\usepackage{cite}
\usepackage{amsmath,amssymb,amsfonts}
\usepackage{algorithmic}
\usepackage{graphicx}
\usepackage{textcomp}
\usepackage{hyperref} 
\usepackage{booktabs} 
\usepackage{multirow}
\usepackage{xcolor}
\usepackage{subcaption} 
\usepackage{lipsum} 
\usepackage{float} 
\usepackage{stfloats} 
\def\BibTeX{{\rm B\kern-.05em{\sc i\kern-.025em b}\kern-.08em
    T\kern-.1667em\lower.7ex\hbox{E}\kern-.125emX}}
\begin{document}

\title{
SRSA: A Cost-Efficient Strategy-Router Search Agent for Real-world Human-Machine Interactions}


\author{\IEEEauthorblockN{1\textsuperscript{st} Yaqi Wang}
\IEEEauthorblockA{\textit{School of Computer Science} \\
\textit{Carnegie Mellon University}\\
Pittsburgh, USA \\
yaqiwang@andrew.cmu.edu}
\and
\IEEEauthorblockN{2\textsuperscript{nd} Haipei Xu}
\IEEEauthorblockA{\textit{Graduate School of Art and Science} \\
\textit{Columbia University}\\
New York City, USA \\
hx2385@columbia.edu}
}

\maketitle

\begin{abstract}
Recently, as Large Language Models (LLMs) have shown impressive emerging capabilities and gained widespread popularity, research on LLM-based search agents has proliferated. In real-world situations, users often input contextual and highly personalized queries to chatbots, challenging LLMs to capture context and generate appropriate answers. However, much of the prior research has not focused specifically on authentic human-machine dialogue scenarios. It also ignores the important balance between response quality and computational cost by forcing all queries to follow the same agent process. To address these gaps, we propose a Strategy-Router Search Agent (SRSA), routing different queries to appropriate search strategies and enabling fine-grained serial searches to obtain high-quality results at a relatively low cost. To evaluate our work, we introduce a new dataset, Contextual Query Enhancement Dataset \href{https://drive.google.com/file/d/1YQj6iuST7YcQ2xPvuUzyHEbDYbcsfwtN/view?usp=sharing}{(CQED)}, comprising contextual queries to simulate authentic and daily interactions between humans and chatbots. Using LLM-based automatic evaluation metrics, we assessed SRSA's performance in terms of informativeness, completeness, novelty, and actionability. To conclude, SRSA provides an approach that resolves the issue of simple serial searches leading to degenerate answers for lengthy and contextual queries, effectively and efficiently parses complex user queries, and generates more comprehensive and informative responses without fine-tuning an LLM.
The code is available at \href{https://anonymous.4open.science/r/SRSA-3A04/}{https://anonymous.4open.science/r/SRSA-3A04/}.
\end{abstract}


\section{Introduction}
As large language models (LLMs) have demonstrated impressive emerging capabilities and gained widespread popularity, researchers have begun leveraging these models to build LLM-based agents. Specifically, they adopt LLM as the main component of the brain or controller of these agents and expand their perception and action space through strategies such as multi-modal perception and tool utilization \cite{xi2023rise}. Meanwhile, the concept of retrieval-augmented generation (RAG) has emerged as a way to address the limitations of LLM, particularly their tendency to "hallucinate" inaccurate information \cite{zhang2023siren} and their difficulty maintaining up-to-date knowledge within its parameters. 

\begin{figure}[t]
\centerline{\includegraphics[width=0.5\textwidth]{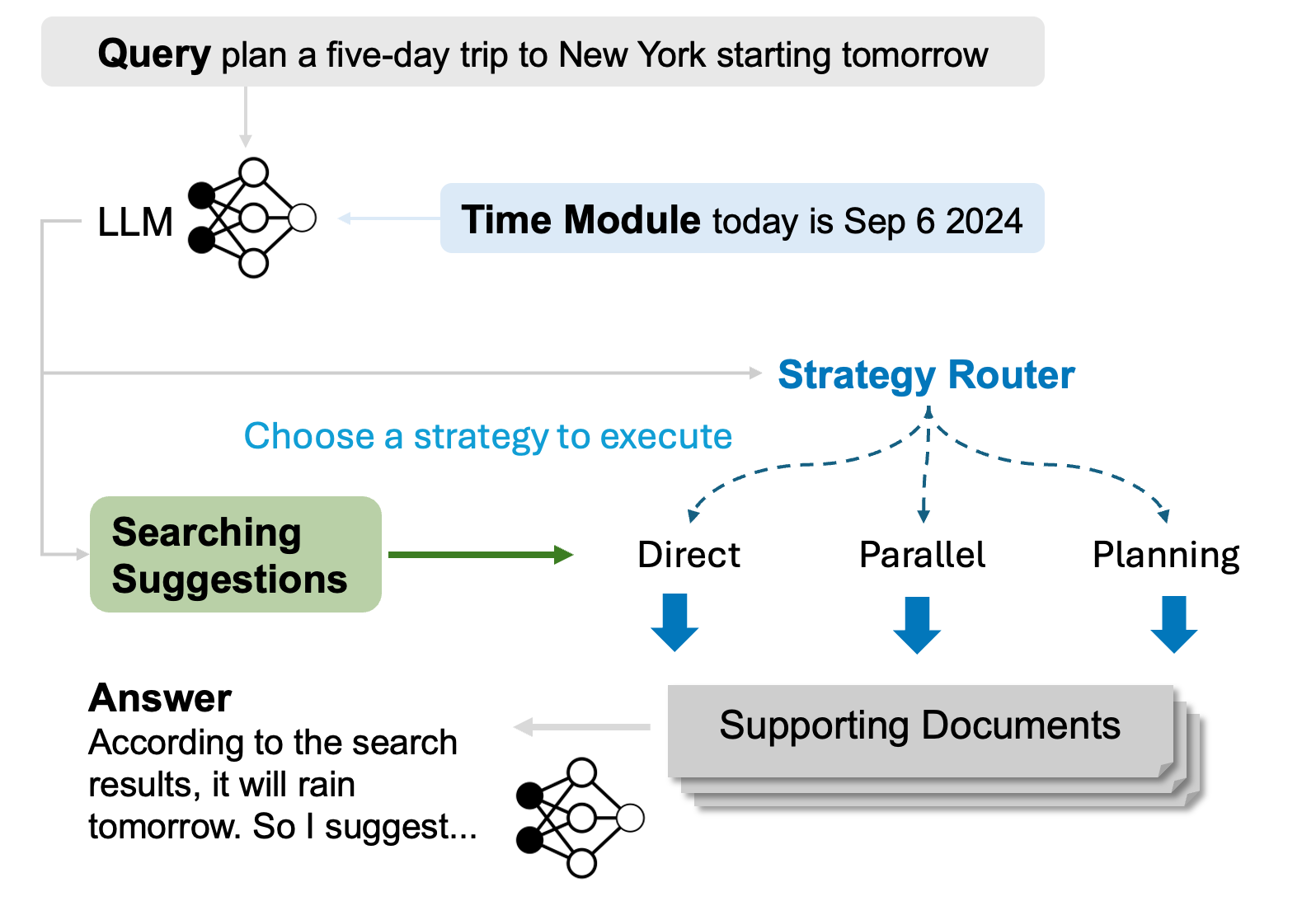}}
\caption{Strategy-Router Search Agent (SRSA) Workflow}
\label{fig:flow}
\end{figure}

Recent works like 'Metacognitive Retrieval-Augmented Large Language Models' \cite{zhou2024metacognitive} created a search agent that has metacognition, achieving better reasoning and planning ability. RA-ISF \cite{RA-ISF} considered self-knowledge before searching and utilized question decomposition to generate reasoning trees, achieving good results in question-answering. ERAGent \cite{shi2024eragent} used enhanced question rewriter and knowledge filter to improve the search agent’s ability in multi-hop question answering. BIDER \cite{jin2024bider} introduced a sophisticated pipeline designed to mitigate inconsistencies between retrieved knowledge and the information required by LLMs, and enhanced the quality of LLM-generated responses while separate training is required for each dataset and generator model, limiting its applicability.

However, these search-agent-related studies lack consideration of the lengthy context and personalized query, where users pose complex questions to chatbots, requiring machines to dynamically adjust their workflow structure based on the context of different questions. For example, a query like 'Plan a three-day trip to New York starting tomorrow' may initially get a result such as 'visit the museum on day one and Central Park on day two.' However, this result overlooks that rain is expected on day two and that a special exhibit is scheduled at the museum on that day. To address such scenarios, a serial search is necessary, where each subsequent query refines the previous one to deliver more relevant and personalized results. For example, the model is expected to start with "What’s the weather for the next three days?" and then follow up with "Are there any specific activities in New York during rainy days?". Moreover, if the initial search results are unsatisfactory, the LLM should identify why and modify the search input accordingly. Could we solve this issue without fine-tuning an LLM? One approach for serial searching is to use the ReAct \cite{yao2022react}, which iterates through 'thought,' 'action,' and 'observation.' However, this often results in too many unrelated search results, which negatively impacts the final answer generation \cite{shi2023large} and makes the process more expensive due to the need for multiple LLM inferences.

Unlike previous studies that primarily focus on simple factual queries, this paper addresses the complexities of real-world user-chatbot interactions. We identify two key challenges in this domain:
\begin{enumerate}
    \item Users frequently pose context-rich and highly personalized questions, often requiring multiple search iterations. The absence of a suitable dataset for such scenarios has hindered progress in this area.
    \item The cost and timeliness of production need to be balanced with quality. Existing approaches \cite{mao2023large, jin2024bider, zhou2024metacognitive, RA-ISF, trivedi2022interleaving} often prioritize result quality at the expense of computational efficiency and response time, leading to costly and potentially impractical solutions for real-time applications.
\end{enumerate}

Therefore, to address the challenges of real-world human-machine interactions, this paper introduces the Strategy-Router Search Agent (SRSA) framework. SRSA innovatively employs three distinct search strategies and an intelligent routing mechanism to automatically direct different queries to the most appropriate strategy. This approach optimizes search results and enhances LLM's output while maintaining computational efficiency, striking a balance between answer quality and cost-effectiveness in authentic dialogue scenarios.

To validate the effectiveness of SRSA, we developed a new data set, Contextual Query Enhancement Dataset (CQED), which focuses on long-context, situational user searches, simulating the complexity of real-world diaglogues. CQED requires agents to demonstrate a nuanced understanding of user requirements and provide useful, non-trivial responses. We evaluated SRSA against two baseline models: an LLM with a single-round search capability and a ReAct-based agent with search functionality. As for the evaluation metrics, we designed four dimensions to evaluate the output of search results: informativeness, completeness, novelty, and actionability. The results demonstrate SRSA's outstanding performance compared to both baseline models, which could generate more informative and complete answers in an efficient way. This paper contributes to improving LLM answer quality in authentic dialogue scenarios, aiming to address a critical gap in current language model applications with a focus on the intricacies of real-world interactions.

\section{Related Work}

\subsection{Prompt Engineering Techniques}

Prompt engineering involves designing and refining the prompts that guide language models to generate specific outputs. This process can significantly enhance the model’s performance on various tasks without modifying the underlying architecture. Methods of prompt engineering include manually crafting prompts that effectively lead the model to desired answers and using automated techniques to optimize prompts based on their performance. Additionally, some advanced strategies involve dynamic prompts that adjust based on the context or through continuous learning. This area is crucial for using pre-trained models in novel applications and improving interaction with LLMs \cite{sahoo2024systematic}. 

Here, we will introduce two techniques that can significantly improve the effect of prompt engineering: Chain of Thought (CoT) \cite{wei2022chain} is a reasoning process in which the model gradually derives a series of intermediate steps or sub-goals before generating the final answer. These intermediate steps form a step-by-step process that ultimately guides the model to the correct result. Moreover, it has recently become an indispensable means to improve the performance of LLM in complex reasoning tasks. In-context learning (ICL) \cite{dong2022survey} has been widely used for LLM prompting, which augments LLM generation by providing a few examples in the prompt. It efficiently benefits the generation quality because it does not need model training or parameter adjustment. 



\subsection{Retrieval Augmented Generation (RAG)}

RAG is a technique that allows language models to capture information from external knowledge to generate better answers. It retrieves relevant information based on the query and then uses it to guide the LLM in developing a response within the retrieved data. According to previous studies, RAG is a powerful tool for improving the accuracy of LLM responses and processing long contexts \cite{lewis2020retrieval}. In our paper, we used RAG to obtain more accurate information, making the LLM's response more comprehensive and high-quality.

\subsection{LLM-based Agents}

With the impressive capabilities of LLMs, researchers have started leveraging them to build AI agents. LLMs serve as the core "brain" or controller of these agents, with strategies like multimodal perception and tool utilization extending their perception and action spaces \cite{xi2023rise}. So, LLM-based Agent is an intelligent entity that treats LLM as a brain. It can use tools (such as search engines and calculators), interact with the environment, and respond and plan what to do next based on the results of the interaction. There are many successful studies on LLM-based agents, such as MetaGPT \cite{hong2023metagpt} featuring the multi-agent framework, Data Interpreter \cite{hong2024data} as a data scientist and Devin, and the AI software engineer \cite{CognitionLabs2024}. 

One of the popular LLM-based Agent frameworks is ReAct (reason+Act) \cite{yao2022react}, which utilizes chain-of-thought reasoning and repeatedly interacts with the environment, thinking and planning the next step based on the results of the interaction. Another one is depth-first search-based decision tree (DFSDT) \cite{qin2023toolllm}. It allows LLM first to generate step-by-step reasoning. When a failure is found, it returns to the state before the reasoning track fails and re-performs reasoning, and the reasoning path is like traversing in a spanning tree with depth-first search method.

\subsubsection{LLM-based Search Agents}
LLM-based search agent utilizes LLM to reason and enable better information retrieval. OpenAI's WebGPT \cite{nakano2021webgpt} fine-tuned a LLM to monitor human search websites to get high-quality answers in long-form question-answering settings.

Query rewriting is a powerful technique for search agents that involves rephrasing, restructuring, or modifying a query. The primary goal of this process is to improve the query's alignment with a search engine. Query rewriting can improve the search quality without adjusting model parameters. Instead of immediately retrieving information using the initial query and generating answers, employing the LLM to rephrase or reformulate the query first will get a better answer \cite{ma2023query}. This method aims to improve the overall effectiveness of LLM-based information retrieval and response generation.
 
    Interleaving Retrieval with Chain-of-Thought (IR-CoT) \cite{wang2023query2doc} is a good work using a multi-step retrieve-and-read approach to solve the multi-step QA problem. Another seminal work in this area is "Query Rewriting for Retrieval-Augmented Large Language Models" \cite{ma2023query}, which demonstrates how rewriting queries can significantly improve the relevance of retrieved information, leading to more accurate and reliable text generation by the model. Query2doc \cite{wang2023query2doc} prompts LLM to generate relevant passages based on the query and then expands the original query by merging the generated passages. Moreover, LLM4CS \cite{mao2023large} rephrases queries based on multi-turn dialogues. It generates many query rewrites and hypothetical responses and then uses aggregation methods to get an integrated representation of the user's search intent. Similarly, InteR \cite{feng2023knowledge} promotes knowledge refinement through multiple rounds of interaction between search engines and LLMs. ERAGent \cite{shi2024eragent} used enhanced question rewriter and knowledge filter to improve the search agent's ability in multi-hop question answering. Another study highlights the use of LLMs as search agents in a trustworthy manner \cite{shi2023know}.

Query rewriting is also important for getting better search results. Usually, there is a semantic gap between the user's queries and documents, and the paraphrasing technique is helpful to find more related documents \cite{wang2013paraphrasing}. A recent work about query expansion \cite{jagerman2023query} used LLM to first generate an answer to the query and rationale and then concatenated the query and model output as the input of the retrieval system.

\section{Methodology}

To better bridge the gap between complex human contextual queries and effective search engine inputs, we developed the dynamic Strategy-Router Search Agent (SRSA) framework.

\begin{figure*}[htbp]
\centerline{\includegraphics[width=0.8\textwidth]{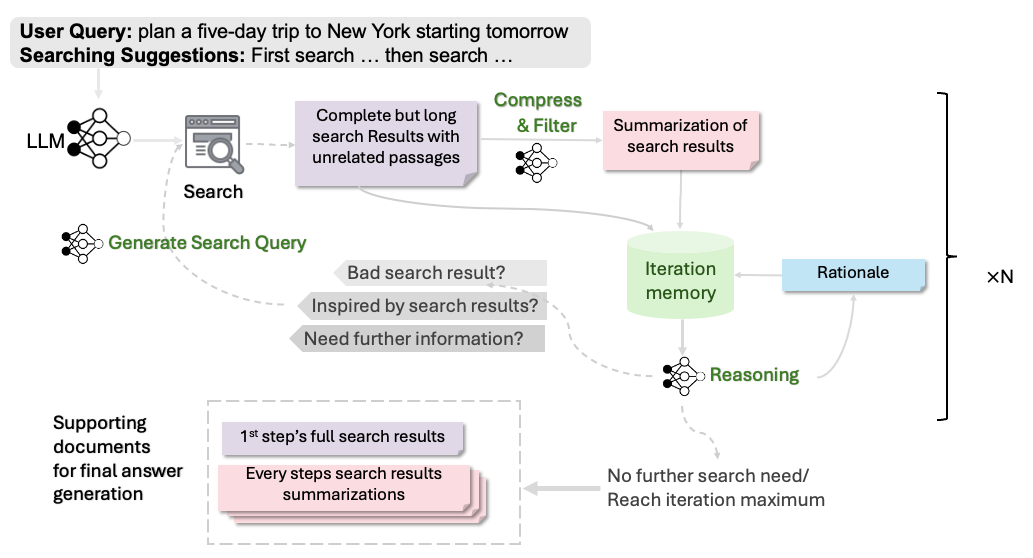}}
\caption{Workflow of planning search strategy module}
\label{fig:planning}
\end{figure*}

\subsection{Framework Overview}

As illustrated in Figure~\ref{fig:flow},the SRSA framework begins by processing the initial user query through an LLM, incorporating a time module to ensure temporal relevance. This step is crucial for handling real-time, context-dependent queries often encountered in authentic human-machine interactions. The LLM then generates two key outputs: The most appropriate search strategy for the query and overview searching suggestions tailored to the chosen strategy.

The core of SRSA, the strategy router, then classifies and routes the query to one of three sophisticated search strategies:
\begin{itemize}
    \item \textit{Direct Search} Rewrites the original query and searches online to retrieve supporting documents. It is designed for simple questions that an LLM can answer effectively with supporting information from a single-round retrieval.
    \item \textit{Parallel Search} Generates and simultaneously pursues multiple related sub-questions, then aggregates and summarizes the results. This strategy excels at handling queries that usually have two or more parallel concepts or tasks.
    \item \textit{Planning Search} Utilizes overview suggestions to plan what to search for at each step and automatically thinks and plans subsequent searches based on the initial results. Queries will be classified into this strategy if the query requires a sequence of searches, where each step's inquiry depends on the information obtained from the previous search.
\end{itemize}

Each search strategy module will not directly produce the final answer to the user's query but will output a detailed supporting documents or reference passage. This reference passage is then used as a prompt to generate the final answer.

Our SRSA framework dynamically balances efficiency and effectiveness. The strategy router plays a crucial role in this balance, intelligently categorizing queries and directing them to the most appropriate search strategy. This approach ensures that simple queries are handled swiftly and efficiently, while complex, multi-faceted questions receive the depth of processing they require.

\subsection{Strategy-Router Module}

Given that $\mathcal{M}(a | b)$ denotes the function using LLM to generate text, where $b$ is the fixed prompt and $a$ are changeable inputs, the search strategy routing prompt $p_{s}$, the user's query $Q$, the chosen strategy $\mathcal{S}$ and the generated suggestions for searching $sug(\mathcal{S})$, the strategy-router module could be formalized as 
\[\mathcal{S}, sug(\mathcal{S}) = \mathcal{M}(Q | p_{s})\]

$\mathcal{S}$ will be one of the following: $\mathcal{D}$ for direct search, $ \mathcal{P}$ for parallel search strategy and $\mathcal{R}$ for planning search strategy. Now, denote $\mathcal{S}(\cdot)$ as the function that takes the search query as input, and outputs the supporting documents for the generation of the final answer. Since there are three search strategies, $\mathcal{S}(\cdot)$ will be one of the following functions: $\mathcal{D}(\cdot), \mathcal{P}(\cdot), \mathcal{R}(\cdot)$.

    \paragraph{Direct Search $\mathcal{D}$}: If a query is classified into direct search strategy, $sug(\mathcal{D})$ can be parsed into a rephrased query $Q_r$, which is more concise and appropriate version of the user's lengthy contextual query for input into the search engine.
    \[Q_r = \text{Parser}(sug(\mathcal{D}))\]
    \[
    \mathcal{D}(Q) = \text{Search}(Q_r)
    \]

    \paragraph{Parallel Search $\mathcal{P}$} This involves generating related sub-questions from the main query and executing these searches in parallel, similar to the question decomposition tasks described in recent works like Least-to-Most \cite{zhou2022least}. The results are then aggregated.

    \[\{s_{par_1}, s_{par_2}, \ldots, s_{par_n}\} = \mathcal{M}(Q_r , sug(\mathcal{P}) | p_{parallel})\] 
    
    This represents the set of generated sub-questions $\{s_{par_m}\}_{m=1}^{n}$ from $Q$, where $p_{parallel}$ is the prompt for the parallel search strategy. The number of sub-questions $n$ is dynamically decided by LLM and is undefined. The parallel search results are a concatenation of all search results of each sub-questions:
    \[
     \mathcal{P}(Q) = \bigcup_{i=1}^{n} \text{Search}(s_{para_i})
    \]
    
    \paragraph{Planning Search $\mathcal{R}$} The general workflow of planning search strategy is shown in Figure~\ref{fig:planning}, similar to IR-CoT \cite{trivedi2022interleaving} but it includes an additional summarization process, selects only a subset of the search results as final reference documents, rather than using all the search results. This module consists of several iterations. In each iteration, the agent compresses the search results from the previous steps (except for the first step) to filter out irrelevant or nonsensical information. Then the agent performs reasoning to: 1) evaluate the quality of the search results; if they are of poor quality, determine why and rewrite the search query; 
2) identify interesting or informative search results from the last iteration and decide whether to explore these points further for deeper insights; 
3) assess whether the search results are sufficient to answer the user's query. If the agent concludes that the current search results stored in iteration memory are enough to provide a comprehensive answer, the iteration will stop. The iteration will also stop if the maximum number of iterations is reached.

Finally, the module outputs the full search results from the first step and the summarization of search results from each step, denoted as $\mathcal{R}(Q)$, as the supporting documents for generating the final answer.

\paragraph{Final Answer Generation}

The final answer is generated based on the original query $Q$, search results reference $\mathcal{S}(Q)$, and a RAG prompt $p_{RAG}$:
\[\text{FinalAnswer} = \mathcal{M} (Q, \mathcal{S}(Q) | p_{RAG})\]

\section{Experiments}
In the experiments section, we begin by introducing the construction of a context-rich question dataset, the Contextual Query Enhancement Dataset(CQED). Subsequently, we describe the setup of our experiments, including the configuration of parameters for our SRSA and details of the prompts, as well as the selection and introduction of baseline models, the choice of search engine API, and the metrics used, as well as how to leverage a LLM for automatic evaluation.

\subsection{Dataset Construction}
We first construct a new dataset that focuses on long-context, user-situated searches, which introduces new challenges to the agent. This task requires a deep understanding of the context and aims to provide users useful and detailed information.

\subsubsection{Why do we need a new dataset}
Some existing datasets for testing Retrieval-Augmented Generation (RAG), such as PopQA \cite{mallen2023llm_memorization}, TriviaQA \cite{2017arXivtriviaqa}, HotpotQA \cite{yang2018hotpotqa}, and the AI2 Reasoning Challenge (ARC) \cite{allenai}, typically provide reference documents along with a deterministic question, such as "What is Henry Feilden's occupation?" These datasets evaluate the model’s RAG ability by testing whether it can extract correct answers from long texts. However, with the current use of LLMs, users often input situational questions rather than simple factual queries.

\begin{figure*}[htbp]
\centerline{\includegraphics[width=0.8\textwidth]{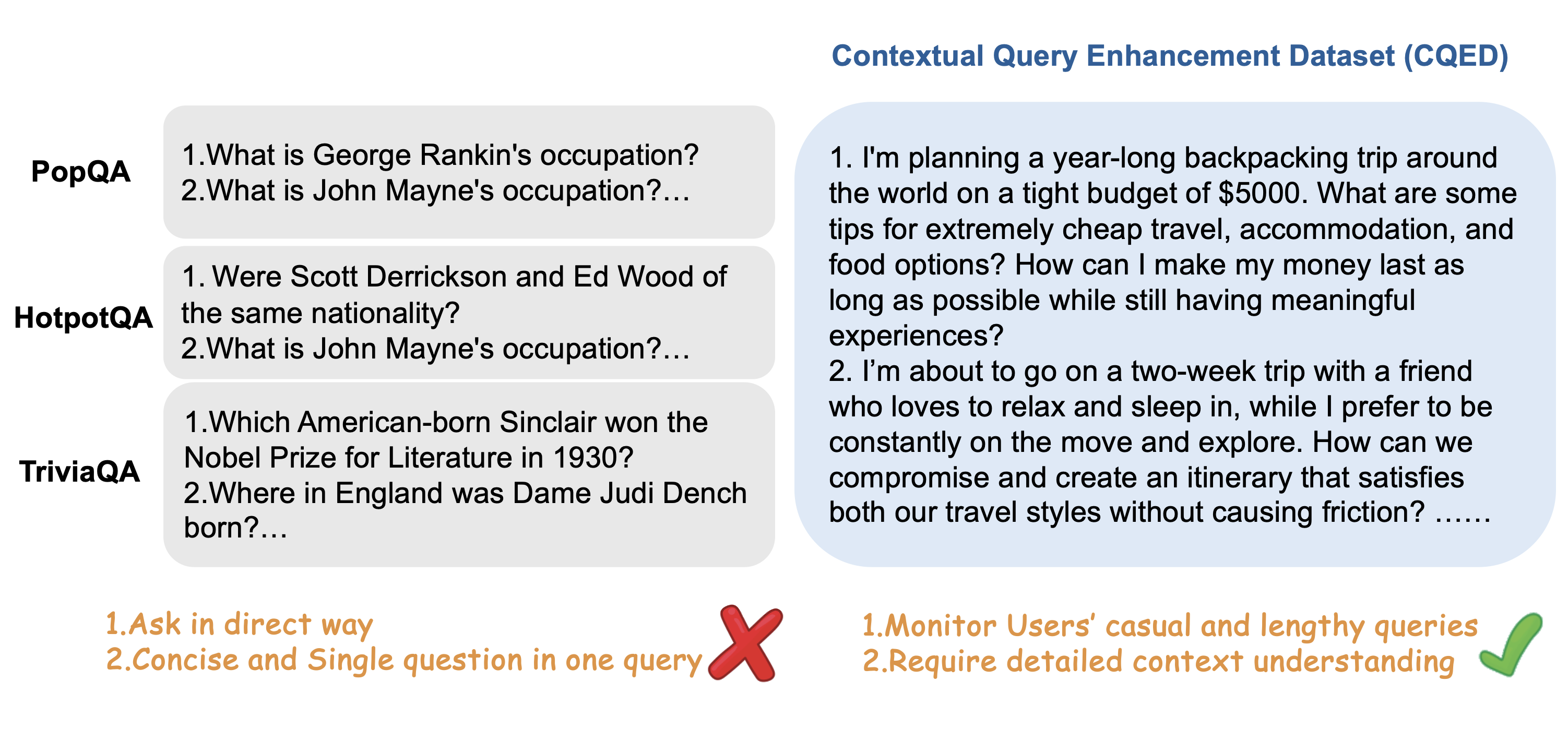}}
\caption{Comparison of the CQED and other QA datasets}
\label{fig:dataset_comp}
\end{figure*}

To further refine the capabilities of our search agent in real-world interactions, we've developed a dataset named "Contextual Query Enhancement Dataset (CQED)." This dataset is crafted to evaluate the agent's ability to navigate complex, user-scenario-based queries that demand high levels of specificity and contextual understanding, as illustrated in Figure~\ref{fig:dataset_comp}. Each query in CQED is designed to simulate real-world situations, such as users seeking detailed product information, travel plans for specific dates or accurate flight schedules. This design necessitates nuanced searches by the agent to effectively assist an LLM in generating accurate responses. 

A distinctive challenge posed by the CQED is the significant semantic divergence between the user's queries and the content typically retrievable through standard search methods. Additionally, since a single question may contain multiple explicit or implicit queries, it is challenging for search agents to generate comprehensive and satisfactory answers.

\subsubsection{Dataset Construction Process}

\begin{figure*}[htbp]
    \centerline{\includegraphics[width=0.7\textwidth]{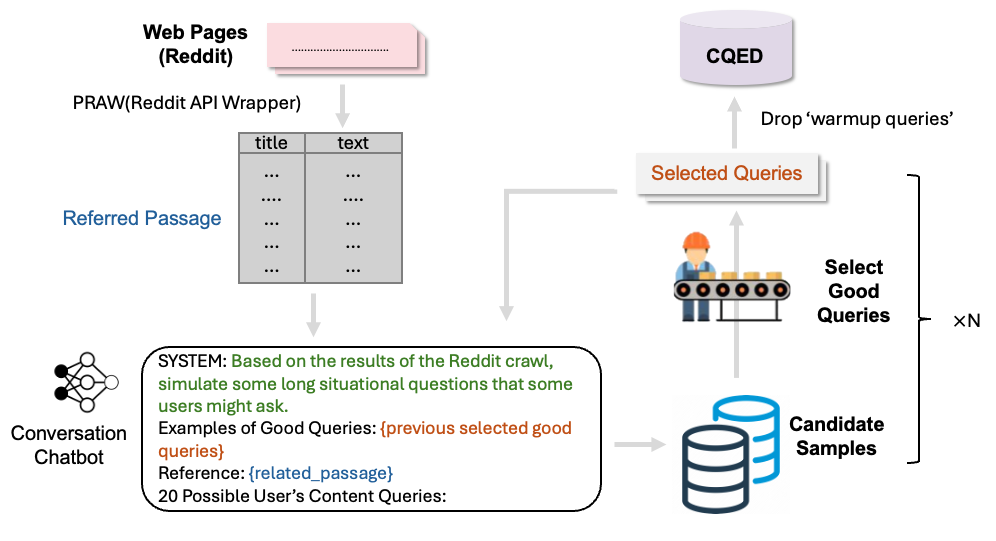}}
    \caption{Construction Process of CQED}
    \label{fig:dataset_process}
\end{figure*}

The construction of the CQED is illustrated in Figure~\ref{fig:dataset_process}. Initially, domains such as shopping, research, travel, and digital devices were established as the primary areas of focus for the dataset. Relevant posts, including titles and content, were retrieved from Reddit using PRAW \cite{praw2024}. 

To construct the dataset, We conduct multiple rounds of conversations with the LLM-based chatbot (Claude 3.5 Sonnet). In each iteration, the conversational LLM-based chatbot is given 100 crawler results. It is then asked to generate 20 scenario-based questions that users might ask. From these, 1 to 3 high-quality answers are manually selected and refined by humans. In the next iteration, previously selected queries will be presented in the prompt so that the chatbot can learn from these selections and modifications. This process is repeated multiple times to generate a total of 182 answers. The answers selected during the initial three iterations are discarded (warm-up queries are removed). Due to the LLM's long context limitation, this cycle is performed independently multiple times.

\subsection{Experiments Setup}

We set up two baseline agents for comparison with our SRSA. One baseline agent uses a single-round search tool, while the other is a ReAct agent with a query rewriting module. The maximum number of ReAct iterations is set to 5, as this has been shown to be a reasonable number in this work \cite{zhou2024metacognitive}. We initially tried setting this number to 3, but it proved insufficient, leading to frequent termination due to reaching the maximum iterations. To ensure reproducibility, the temperature for LLMs is set to 0, \texttt{max\_token} equals the context window length of the model, and \texttt{n = 1} (the number of responses generated for each input message) to minimize costs.

For the pre-trained LLMs used in the experiments, we selected Google’s gemma-2-2b-it \cite{gemma_2024}, Meta’s Meta-Llama-3-8B-Instruct \cite{llama3modelcard}, and Mistral-7B-Instruct-v0.3 \cite{mistral7b_2023}. For evaluation, we employed GPT-4o-mini as the judging model, presenting it with the answers from the three search agents simultaneously. This allows the judging model to compare the results of our SRSA with those of the baseline agents.

\subsubsection{Baseline Models}
There are two baseline agents used for comparison:
    The first baseline agent is a single-round search Agent, referred to as the "simple search agent" in the following sections. This baseline agent performs a one-time search, and the search result is used to generate the final output. The second baseline agent is a ReAct-based search agent that integrates the ReAct flow (Thought, Action, Observation) with the search tool. It includes a rephrasing module that reformulates the contextual query into a precise and clear question for the search engine. The rephrasing module is incorporated into the ReAct-based search agent for a better comparison, as our SRSA has a similar processing component. The maximum iteration times are set to be 5. Generally, after receiving a query, the ReAct-based search agent analyzes and thinks about what to search for (Thought step), then searches (using the Search Engine API, discussed in the next section), and then proceeds to the next thought step (Next Thought step) based on the result (Observation). There is no summarization process here, and the final results are generated based on the entire iteration history.


\subsubsection{Search Engine API}
LLM has to use a tool for an online search to enable the RAG ability. Here, we choose Tavily Search API \cite{tavily2024api}. Tavily Search API is a search engine optimized for LLMs and RAG that is aimed at efficient, quick, and persistent search results. Unlike other search APIs such as SerpAPI \cite{serpapi} or Google Custom Search API \cite{googlesearchapi}, Tavily focuses on optimizing search for AI developers and autonomous AI agents. The parameters for the Tavily Search API are: \texttt{depth = advanced}, \texttt{domain = general} and \texttt{max\_results = 5}.

The system possesses several merits that enhance its functionality and user experience. Firstly, it employs a simple REST API call method for operation. Unlike some systems, it does not merely output a snippet from the relevant website, which often results in incomplete information. Additionally, it integrates several key processes, such as searching, scraping, filtering, re-ranking, and extracting the most relevant information, which optimizes the accuracy and relevance of the results. Moreover, the Tavily Search API allows users to specify the search domain. The 'news' domain is specifically tailored for searching the latest news and restricts the search to well-known and credible news release websites, ensuring the reliability of the information retrieved. In contrast, the 'general' domain imposes no such restrictions, offering broader search capabilities \cite{tavilydocs}.

The unified third-party API, Tavily Search API, can reduce the manual work of crawling, cleaning, and re-sorting web pages and focus more on the design of the search agent's workflow itself. In addition, all search tools use the same third-party API to ensure the fairness and rationality of the experiment.

\subsubsection{Evaluation Metrics}
Studies have shown that LLM can be used to automatically evaluate text robustly \cite{chiang2023closer, chiang2023can}. We use the LLM as an automatic evaluation machine, with our evaluation metrics including:
\begin{enumerate}
    \item Informativeness (0-5 points): Measures the degree of information richness in the answer and the amount of useful information provided. Higher scores are awarded to answers that include a greater proportion of relevant content valuable to users.
    
    \item Completeness (0-5 points): Assesses how well the answer addresses all aspects of the user's question. This is evaluated by breaking down the query into its constituent concepts and checking if the answer covers all these concepts.
    
    \item Novelty (0-5 points): Evaluates the extent to which the answer provides information that requires searching and is not common knowledge. Answers that include less obvious or difficult-to-obtain information score higher.
    
    \item Actionability (0-5 points): Measures the extent to which the user can take specific actions based on the answer. More specific and actionable answers receive higher scores, while abstract or vague answers score lower.
\end{enumerate}

Each metric is scored on a scale of 0-5, with explanations provided to justify the scores. Evaluators reference specific parts of the answers or summarize them to support their scoring decisions. The scoring process involves a comparison between different agent types (simple search, ReAct-based search, and SRSA) to ensure relative performance is accurately reflected. 

To ensure that the results of the automatic LLM evaluation met our expectations, we manually evaluated 8 data sets. However, each example is very long. Using too many examples in one prompt will confuse the referee model, and the middle examples may be forgotten as in the case of long prompts, the middle prompt gets \cite{LostInMiddle}. To address this, we conducted 2 evaluations per data point, using 2 4-shot prompts for each evaluation, so all examples in our manual evaluation were utilized.

To obtain compelling results, we implemented t-test to determine whether there is a statistically significant difference between the two models. We calculated the sample means and sample variances, then derived the t-statistic and p-value, assuming that the results of the automatic evaluation follow a truncated normal distribution. The t-stat is defined as
\[
t = \frac{\bar{X}_1 - \bar{X}_2}{\sqrt{\frac{s_1^2}{n_1} + \frac{s_2^2}{n_2}}}
\]
where $\bar{X}_1$ and $\bar{X}_2$ are the sample means of the two groups, $s_1^2$ and $s_2^2$ are the sample variances of the two groups,  and $n_1$ and $n_2$ are the sample sizes of the two groups.

\subsection{Experiments Results}


According to the previous section, we used three LLM models, which we now refer to as Gemma, Llama, and Mistral. Unfortunately, during our experiments, we found that Gemma and Llama were unable to effectively follow formatted output based on the system prompt. Our SRSA requires LLMs to have strong formatted output capabilities, allowing the program to parse and advance the process based on the LLM's formatted output. The ReAct-based search agent also has similar requirements for LLMs. When executing the strategy router module, Gemma and Llama were unable to output which class of strategy they chose in a formatted manner, so they were automatically categorized as Direct. However, this doesn't mean that the test data from these two models is useless. We can compare the results of Gemma and Llama running SRSA when choosing 'Direct' as a targeted strategy with the results of a simple search agent, demonstrating the importance of rephrasing in situations where lengthy queries require a search.

Fortunately, our Mistral model was able to follow the output of our SRSA model well, so our conclusions are primarily based on the results from this Mistral model.

\subsubsection{Power of Rephrasing}

\begin{figure*}[htb] 
    \centering
    \begin{subfigure}[b]{0.48\textwidth}
        \centering
        \includegraphics[width=\textwidth]{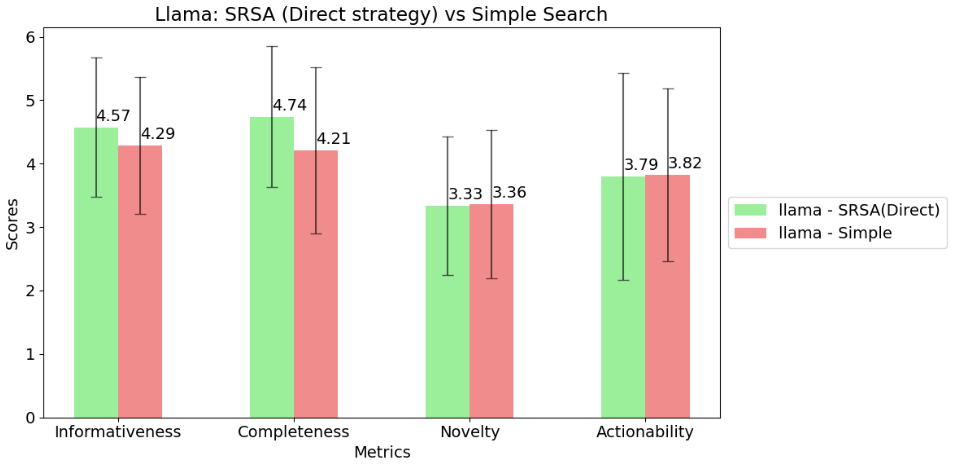} 
        \caption{Compare the performance of SRSA(Direct) and simple search agent on samples where Llama-based SRSA choose to use Direct search strategy. The numbers are the average scores, and the error bar is the 95\% confidence interval (same for other bar charts).}
        \label{fig:subfig1}
    \end{subfigure}
    \hfill
    \begin{subfigure}[b]{0.48\textwidth}
        \centering
        \includegraphics[width=\textwidth]{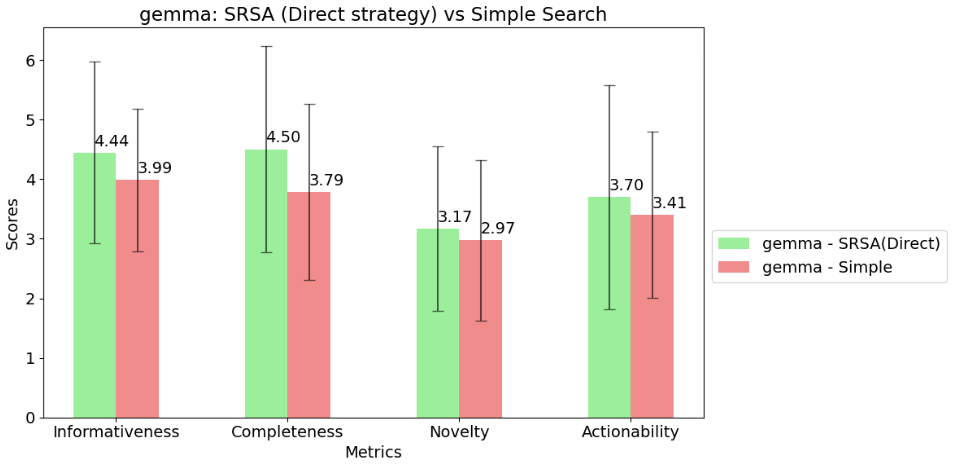} 
        \caption{Compare the performance of SRSA(Direct) and simple search agent on samples where gemma-based SRSA choose to use Direct search strategy.}
        \label{fig:subfig2}
    \end{subfigure}
    \caption{Compare Llama and gemma's performance on SRSA(Direct) and simple search agent, indicating that rephrasing before searching could reduce the gap between human contextual questions and the query put into the search engine.}
    \label{fig:ex0}
\end{figure*}

We can assess the capabilities of the rephrasing module by comparing the results of SRSA's direct search strategy with those of the simple search agent because the only difference is that SRSA's direct search strategy rewrites the query before doing a search. As shown in Figure~\ref{fig:ex0}, the Llama-based SRSA, which defaulted to the direct search strategy due to its difficulty in properly following formatting requirements, still significantly outperforms the simple search agent in terms of informativeness and completeness. The statistical analysis shows t-statistics of 3.6302 and 6.3128, with p-values of 0.0003 and 0.0000, respectively. Similarly, the Gemma-based SRSA using the Direct strategy also shows significantly higher scores in informativeness and completeness compared to the simple search agent, with t-statistics of 4.2606 and 5.9207, and p-values of 0.0000 and 0.0000.

Although these two LLMs have not fully exploited the advanced routing capabilities of SRSA, the results suggest that even basic rephrasing offers substantial benefits. The data indicates that rephrasing is crucial for handling lengthy and complex queries, as precise reformulation improves the retrieval of relevant information. Since user inputs are often lengthy and varied, effective rephrasing is necessary to generate queries that are more suitable for search engines.

\begin{figure*}[htbp]
    \centerline{\includegraphics[width=0.8\textwidth]{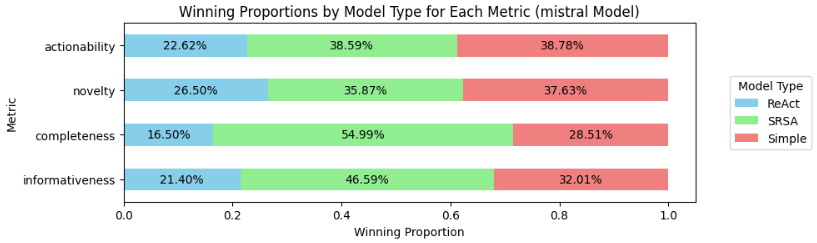}}
    \caption{Wining Proportions of different agents. The naive ReAct-based agent has the lowest winning rate in all metrics, lower than the single-round search agent, indicating a degeneration phenomenon. }
    \label{fig:ex1-3}
\end{figure*}

\subsubsection{Degeneration of ReAct Search Agent for Contextual Queries}

\begin{table}[htbp]
\centering
\caption{T-test for Simple vs ReAct Search Agents.}
\label{tab:reAct_degeneration}
\begin{tabular}{|l|c|c|}
\hline
\textbf{Metric} & \textbf{t-stat} & \textbf{p-value} \\
\hline
Informativeness & 10.8144 & 0.0000* \\
\hline
Completeness    & 11.9862 & 0.0000* \\
\hline
Novelty         & 10.1833 & 0.0000* \\
\hline
Actionability   & 10.9479 & 0.0000* \\
\hline
\end{tabular}
\end{table}

The performance of the ReAct search agent in various scenarios is illustrated. We analyzed the win rate of each agent when answering the same question, determining the proportion of times an agent outperformed others. In cases where two agents tied for first place, both were counted as winners. The win rates were then normalized to 1, resulting in the data shown in Figure~\ref{fig:ex1-3}. It is clear that the ReAct-based agent consistently shows lower performance across all metrics compared to the simpler single-round search agent, indicating significant performance degradation. To strengthen the findings, we conducted a t-test to assess whether the differences in mean scores between the simple search agent and the ReAct-based agent are statistically significant. As shown in Table~\ref{tab:reAct_degeneration}, the t-test results reveal that the ReAct-based search agent significantly underperforms relative to the simple search agent.

This finding suggests that ReAct's iterative querying process struggles when handling complex, context-rich queries that require long and serial search steps. The results imply that as query complexity increases, ReAct's ability to extract relevant information diminishes, leading to less reasonable responses. This is because, during the multiple search steps in the ReAct workflow, the results obtained at each stage, such as web snippets, often end up being either only loosely related to the user’s query or irrelevant. These irrelative documents defect the generated answers, as discussed in this work \cite{shi2023large}. This is the inspiration for our design of the 'Planning' search strategy in SRSA. Each search requires a compressing and filtering process to ensure that when the final answer is generated, the reference documents provided to LLM are highly relevant.

\subsubsection{Comparison between Baselines}

Figure~\ref{fig:ex1-3} and Figure~\ref{fig:ex2-3} highlight the strength of the search router; the two figures are all based on the Mistral model. As shown in Figure~\ref{fig:SRSA-other-agents}, the Mistral-based SRSA significantly outperforms both the ReAct agent and the simple search agent in terms of informativeness and completeness, with t-statistics of 14.9827 and 4.8846, respectively, and p-values of $0.00$ for both. 

However, there are two metrics, novelty and actionability, where SRSA does not demonstrate a significant improvement; instead, it performs similarly to the simple search agent. In terms of novelty, this may be because performing more searches does not significantly enhance novelty. A single-round search can still uncover information beyond common knowledge, and the simple search agent generates answers directly from the search results, which often include "information that human users may not know or find difficult to think of without searching." As for actionability, while the simple search agent can provide actionable recommendations, it may lack completeness. For instance, the suggestions might be incomplete or insufficient to fully address all of the user's concerns or expectations.

 Overall, by dynamically selecting the optimal strategy based on query characteristics, the search router enables SRSA to outperform other agents. In particular, it makes the answers more complete, which means it will answer all the users' needs. Besides, the proportion of helpful information in the answers is also higher. 

\subsubsection{Power of Search Router}
\begin{figure*}[htb] 
    \centering
    \begin{subfigure}[b]{0.48\textwidth}
        \centering
        \includegraphics[width=\textwidth]{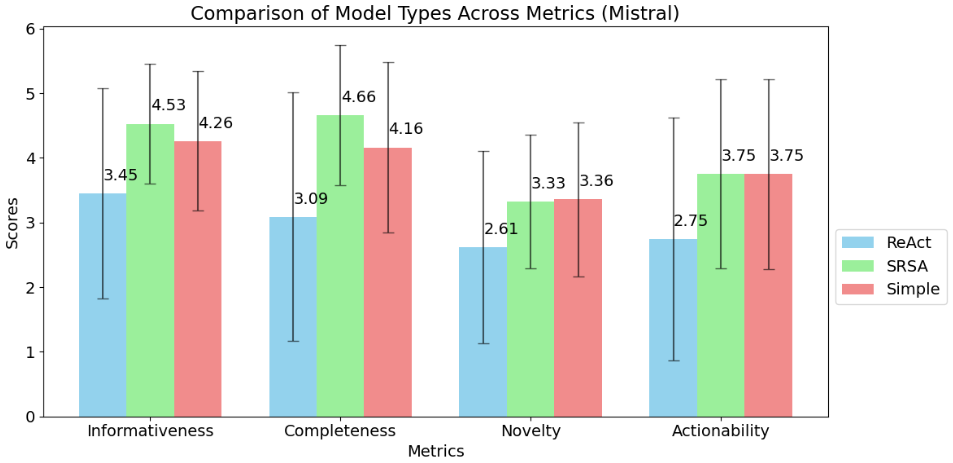} 
        \caption{Compare the performance of Mistral-based SRSA and other agents (the simple search agent and the ReAct-based search agent). Only Mistral's results are considered. SRSA significantly performs better than ReAct agent and simple search agent in informativeness and completeness, as shown in (a) with t-stat equals 14.9827 and 4.8846, p-value = $0.00$ and $0.00$.}
        \label{fig:SRSA-other-agents}
    \end{subfigure}
    \hfill
    \begin{subfigure}[b]{0.48\textwidth}
        \centering
        \includegraphics[width=\textwidth]{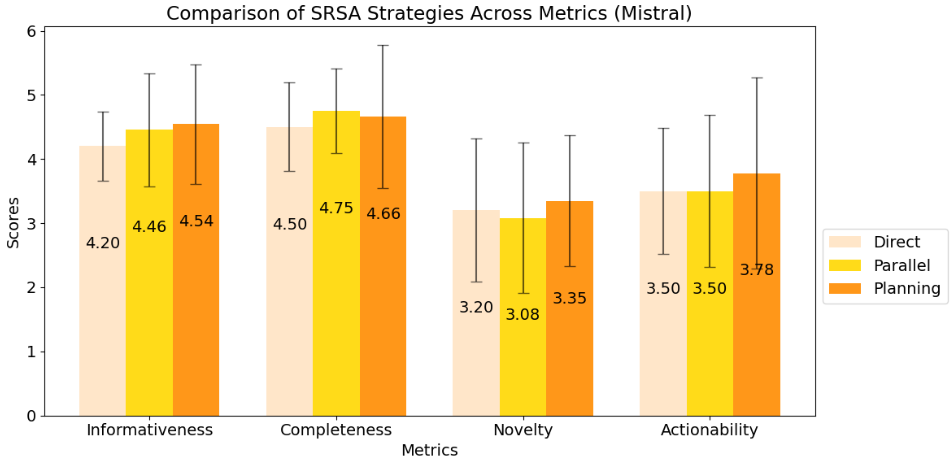}
        \caption{Compare performance of different search strategies of SRSA. the most fine-grained strategy, 'Planning,' tops all metrics except completeness, which 'Parallel' strategy is better at.}
        \label{fig:SRSA-strategies}
    \end{subfigure}
    \caption{Comparison SRSA to baseline models and different search strategies' performance}
    \label{fig:ex2-3}
\end{figure*}

Figure~\ref{fig:SRSA-strategies} offers a closer look at the specific strategies employed by the search router. The planning strategy emerges as the most powerful in terms of informativeness and other metrics, except completeness, where the parallel strategy performs better. This suggests that when dealing with complex, multi-step queries, the planning strategy excels at ensuring detailed and accurate responses. In contrast, the parallel strategy's advantage in completeness indicates its ability to capture a broader range of relevant information in less complex queries. 


It is important to note that our initial intention with the strategy router was to categorize each query into its appropriate strategy, thereby reducing unnecessary LLM inference while maintaining the quality of results. To validate the capability of the search router, we need to verify that the answers to queries assigned to different strategies do not significantly differ in quality. Additionally, we aim to demonstrate that without the search router, when all questions are processed through a single strategy, there are significant differences in the quality of query answers.

Table~\ref{table:sr-comparison} presents a comparison of the direct, parallel, and planning strategies across different metrics. The results show that there are few significant differences between strategies with the search router, as indicated by the p-values. Only the comparison between Direct and Planning strategies for Informativeness shows a significant difference ($p < 0.05$). These findings support the effectiveness of the search router in appropriately assigning queries to strategies, thereby maintaining consistent quality across different approaches while reducing computational cost.

\begin{table*}[h]
\centering
\small
\begin{tabular}{|l|l|c|c|}
\hline
\multirow{2}{*}{\textbf{Comparison}} & \multirow{2}{*}{\textbf{Metric}} & \multicolumn{2}{c|}{\textbf{w/ Search Router}} \\
\cline{3-4}
& & \textbf{t-stat} & \textbf{p-value} \\
\hline
\multirow{2}{*}{Direct vs. Parallel} 
& Informativeness & -1.4467 & 0.1728 \\
& Completeness & -1.3465 & 0.2188 \\
\hline
\multirow{2}{*}{Direct vs. Planning} 
& Informativeness & -2.6741 & 0.0461* \\
& Completeness & -0.9694 & 0.3798 \\
\hline
\multirow{2}{*}{Parallel vs. Planning} 
& Informativeness & -0.6235 & 0.5438 \\
& Completeness & 0.8450 & 0.4104 \\
\hline
\end{tabular}
\caption{After searching the router, there is no significant difference in the results obtained by different strategies.}
\label{table:sr-comparison}
\end{table*}

Furthermore, by correctly assigning different queries to suitable strategies, the model achieves consistent performance across queries categorized under different strategies. This means that classification into the direct strategy is sufficient for certain queries to obtain a good answer, eliminating the need for more complex strategies, like the planning strategy. Our search router successfully identifies these queries, assigning them to simpler strategies, thereby reducing cost. 

This approach demonstrates the search router's ability to balance between response quality and computational efficiency. By intelligently routing queries to the most appropriate strategy, SRSA can provide high-quality answers while controlling the computation cost. This optimization is particularly valuable in real-world applications where both response quality and system efficiency are crucial.

\section{Limitation}


In constructing the dataset, considerations regarding the legality and reliability of data sources were paramount. The data was extracted using PRAW from Reddit, raising potential concerns about copyright and data privacy. Besides, the dataset size was limited since manual screening takes too much time and money. Moreover, there is no perfect method to evaluate the quality of a dataset; it is only manual evaluation, which is very subjective. Moreover, in our experiments, only one LLM model (Mistral) successfully implemented the full SRSA framework, which may affect the generalizability of the results. Besides, the number of baselines might not be enough; other reasoning frameworks like IR-CoT \cite{trivedi2022interleaving} might be added to the baselines to get more convincing results.

\section{Conclusion}

We created a context-rich scenario dataset, CQED, to simulate real human queries and introduce a Strategy-Router Search Agent (SRSA) that enhances the informativeness, completeness, and applicability of responses while utilizing a frozen large language model. By dynamically routing queries of different complexity into appropriate search strategies, SRSA not only improves the quality of answers but also optimizes inference time, increasing efficiency. Our search agent outperforms the basic single-round search agent and addresses the degeneration issues observed in the ReAct-based search agent, which underperforms compared to a single-round agent. In conclusion, this paper introduced the SRSA to deliver high-quality responses in real-world, context-rich, and personalized human-chatbot interactions at a relatively low computational cost.

\end{document}